\documentclass[10pt,twocolumn,letterpaper]{article}
\pdfoutput=1

\usepackage{cvpr}
\usepackage{times}
\usepackage{epsfig}
\usepackage{graphicx}
\usepackage{amsmath}
\usepackage{amssymb}

\usepackage[numbers]{natbib}
\usepackage{stmaryrd}
\usepackage{makecell}
\usepackage{multirow}
\usepackage{sg-macros}

\usepackage[breaklinks=true,bookmarks=false]{hyperref}

\cvprfinalcopy 

\def\cvprPaperID{1565} 

\ifcvprfinal\fi
\begin{document}

\title{DeepPermNet: Visual Permutation Learning}

\author{Rodrigo Santa Cruz\quad Basura Fernando\quad Anoop Cherian\quad Stephen Gould\\
Australian Centre for Robotic Vision, Australian National University, Canberra, Australia\\
{\tt\small firstname.lastname@anu.edu.au}
}

\maketitle
\thispagestyle{empty}

\begin{abstract} 
We present a principled approach to uncover the structure of visual data 
by solving a novel deep learning task coined \emph{visual permutation learning}. 
The goal of this task is to find the permutation that recovers the structure of data from shuffled versions of it. 
In the case of natural images, this task boils down to recovering the original image from patches shuffled by an unknown permutation matrix. 
Unfortunately, permutation matrices are discrete, thereby posing difficulties for gradient-based methods. 
To this end, we resort to a continuous approximation of these matrices using doubly-stochastic matrices which we generate from standard CNN predictions using Sinkhorn iterations. 
Unrolling these iterations in a Sinkhorn network layer, we propose~\emph{DeepPermNet}, an end-to-end CNN model for this task.

The utility of DeepPermNet is demonstrated on two challenging computer vision problems, namely, (i) relative attributes learning and (ii) self-supervised representation learning. Our results show state-of-the-art performance on the Public Figures and OSR benchmarks for (i) and on the classification and segmentation tasks on the PASCAL VOC dataset for (ii).

\end{abstract}

\section{Introduction}
Visual data encompasses rich spatial (and temporal) structure, which is often useful for solving a variety of computer vision problems. For instance, surrounding background usually offers strong cues for object recognition, sky and ground usually appear at predictable locations in a scene, and objects are made up of known parts at familiar relative locations.
Such structural information within visual data has been used to improve inference in several problems, such as object detection and semantic segmentation \citep{Mottaghi:CVPR14,saxena2009make3d,marszalek2009actions}. 

\begin{figure}[t]
	\begin{center}		
		\includegraphics[width=0.47\textwidth]{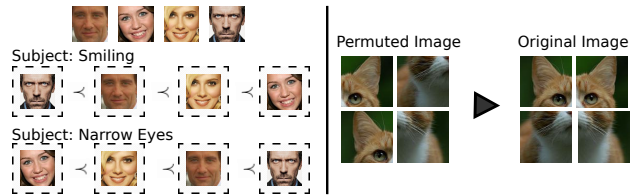}		
	\end{center}
	\label{fig:perm_intro}
	\caption{Illustration of the proposed permutation learning task. The goal of our method is to jointly learn visual features and the predictors to solve the visual permutation problem. This can be applied to ordering image sequences (left) or recovering spatial layout (right).}
\end{figure} 
In this paper, we present a deep learning framework that uses the inherent structure in data to solve several visual tasks. As an example, consider the task of assigning a meaningful order (with respect to some attribute) to the images shown in the left panel of \figref{fig:perm_intro}.
Indeed, it is difficult to solve this task by just processing a single image or even a pair of images at a time.
The task becomes feasible, however, if one exploits the structure and the broader context by considering the entire set of images jointly.
Then, we start to recognize shared patterns that could guide the algorithm towards a solution. A similar task involves recovering the spatial structure in images. For example, consider the task shown in the right panel of \figref{fig:perm_intro}. Here we ask the question ``\emph{given shuffled image patches, can we recover the original image?}". Although this is a difficult task (even for a humans), it becomes easy once we identify the object in the patches (e.g., a cat), and arrange the patches for the recognized object, thereby recovering the original image. 

Such shuffled images can be generated cheaply and in abundance from natural images. 
The problem of recovering the original image from shuffled ones can be cast in an unsupervised learning setting, similar to autoencoder methods~\cite{bengio2007scaling} popular in the neural networks literature. Here the recovery task does not require any human annotations (and is thus unbiased~\citep{Torralba:CVPR11}). Instead it uses the spatial structure as a supervisory signal. Such a learning task is popularly known as self-supervised learning~\cite{Doersch:ICCV15,Fernando2017,Misra:2016,Noroozi:ECCV16}, and is very useful to learn rich features, especially in the context of training deep learning models, which often require large amounts of human annotated datasets. 

As is clear, the aforementioned tasks are similar and essentially involve learning a function that can recover the order, i.e., infer the shuffling permutation matrix. Such an inference could be used to understand scene structure, visual attributes, or semantic concepts in images or image regions. The knowledge acquired by this sort of learning framework can then be used to solve many other computer vision tasks, such as learning-to-rank \citep{Fernando:ICCV2015}, image reconstruction \citep{Brown:2008} and object segmentation \citep{Mottaghi:CVPR14}. 

In this paper, we address the problem of learning to predict visual permutations. Towards this end, we propose a novel permutation prediction formulation and a model based on convolutional neural networks that can be trained end-to-end. Moreover, our formulation admits an efficient solution and allows our method to be applied to a range of important computer vision problems including relative attribute learning and representation learning.

Our contributions are threefold. First, we propose the \emph{Visual Permutation Learning} task as a generic formulation to learn structural concepts intrinsic to natural images and ordered image sequences. Second, we propose the \emph{DeepPermNet} model, a end-to-end learning framework to solve the visual permutation problem using convolutional neural networks. Last, we introduce the \emph{Sinkhorn layer} that transforms standard CNN predictions into doubly-stochastic matrices using Sinkhorn iterations; these matrices are continuous approximations to discrete permutation matrices, and thus allow efficient learning via backpropagation.

We evaluate our proposed approach on two different applications: relative attributes and self-supervised representation learning. More specifically, we show how to apply our method to accurately and efficiently solve the relative attributes task. Furthermore, we show how to learn features in a self-supervised manner achieving the best performance over existing approaches on both object classification and segmentation.

\section{Related Work}
Broadly speaking, the permutation learning problem consists of learning a meaningful order for a collection of images or image regions based on some predetermined criterion. Variations of this task have been studied extensively by many scientific communities. 

In computer graphics, the jigsaw puzzle problem consists of reconstructing an image from a set of puzzle parts \citep{Cho:PMAI10,Sholomon:CVPR13}. Likewise, structured problems including DNA or RNA modeling in biology \citep{Marande:2007} and re-assembling relics in archeology \citep{Brown:2008}, can be modeled as permutation learning problems. We propose a generic data-driven approach for learning permutations. We also develop a CNN based framework to efficiently solve such a problem, which can be applied in diverse applications, although in this paper we limit our scope to computer vision, and review below topics that are most similar to the applications considered in the sequel.

\noindent\textbf{Visual Attributes.} Visual attributes are human understandable visual properties shared among images. They may range from simple visual features (such as ``narrow eyes'' and ``bushy eyebrows'' in faces) to semantic concepts (like ``natural'' and ``urban'' scenes), or subjective concepts (such as ``memorability'' and ``interestingness'' of images). Due to the expressiveness of visual attributes, researchers have successfully used them for many applications, including image search \citep{Kovashka:CVPR12}, fine-grained recognition \citep{Branson:CVPR13} and zero-shot learning \citep{Parikh:CVPR11,Lampert:CVPR14}.

Visual attributes are traditionally treated as binary predicates indicating the presence or absence of certain properties in an image. From this perspective, most of the existing methods use supervised machine learning, whose goal is to provide mid-level cues for object and scene recognition \cite{Farhadi:ECCV10}, or to perform zero-shot transfer \cite{Lampert:CVPR14}. However, there are methods that can discover binary visual attributes in an unsupervised way \citep{Shankar2015,Huang2016}.

A more natural view on visual attributes is to measure their strength in visual entities. For instance, \citet{Parikh:CVPR11} introduced the problem of relative attributes, in which pairs of visual entities are compared with respect to their relative strength for any specific attribute. This problem is usually cast as a learning-to-rank problem using pair-wise constraints. Following this idea, \citet{Parikh:CVPR11} propose a linear relative comparison function based on the well-known Rank-SVM \citep{Joachims:SIGKDD06}, while~\citet{Yu:CVPR14} uses a local learning strategy. 

With the recent success of deep learning and end-to-end learning methods in computer vision, CNN-based methods to tackle the relative attributes problem have been developed. \citet{Souri:ACCV16} jointly learns image representation and ranking network to perform pair-wise comparisons according to a certain attribute. Similarly, \citet{Singh:ECCV2016} propose to combine spatial transformer networks \citep{Jaderberg:NIPS15} and rank networks to localize, in addition to compare visual attributes. Differently from our proposed approach, the aforementioned methods use only pair-wise relationships and do not leverage structure within longer image sequences.
  
\noindent\textbf{Self-Supervised Representation Learning.} The main idea of self-supervision is to exploit supervisory signals, intrinsically in the data, to guide the learning process. In this learning paradigm, a model is trained on an auxiliary task that provides an intermediate representation which can be used as generic features in other tasks. In the deep learning domain, this approach is well-suited as a pre-training procedure in situations when there is insufficient data to support fully supervised learning. 

Towards this end,~\citet{Doersch:ICCV15} uses spatial co-location of patches in images,~\citet{Wang:ICCV15} uses object tracking in videos to provide similar representations for corresponding objects,~\citet{Agrawal:ICCV15} uses labellings produced by ego-motion sensors,~\citet{Fernando2017} uses odd-one-out question answering, and~\citet{Pathak:2016} explores image context to recover missing parts in an image. The only objective in all these methods is to learn visual representations, whereas our proposed method can be used to solve a broader set of problems.

On the other hand, some pretext tasks can be useful themselves. \citet{Isola:2015} learns to group visual entities based on their frequency of co-occurrence in space and time.~\citet{Zhang:ECCV2016} proposes a model to provide plausible color versions for grayscale images.~\citet{Donahue:2016} builds a generative model for natural images.  Note, however that, these methods are highly engineered for their training task and they can not be easily extended to deal with other applications.

A recent work closely related to ours is~\citet{Noroozi:ECCV16} that also proposes to train CNNs for solving image-based jigsaw puzzles. However, different from us, they train a CNN to predict only a tiny subset of possible permutations generated from an image shuffling grid of size $3\times 3$ (specifically, they use only 100 permutations from 362k possible permutations). Instead, our method can handle the full set of permutations and is scalable to even finer shuffling grids. In addition, our scheme is generic and can be used to solve problems such as relative attributes, learning-to-rank, and self-supervised representation learning, and can explore the structure of image sequences or spatial layout of image regions.

\section{Learning Visual Permutations}
In this section, we describe our method for learning visual permutations. We start by formalizing the visual permutation prediction task. Then, we describe our end-to-end learning algorithm, CNN model, and inference procedure.

\subsection{Task}
Given a sequence of images ordered by a pre-decided visual criterion, we generate shuffled sequences by applying randomly generated permutation matrices to the original sequences. Similarly, we can recover the original sequences from the shuffled ones by ``un-permuting" them using the inverse of the respective permutation matrices. In this context, we define the \emph{visual permutation learning} task as one that takes as input a permuted sequence and produces as output the permutation matrix that shuffled the original sequence.

Formally, let us define $X$ to be an ordered sequence of $l$ images in which the order explicitly encodes the strength of some predetermined criterion $c$. For example, $c$ may be the degree of ``smilingness'' in each image. In addition, consider an artificially permuted version $\tilde{X}$ where the images in the sequence $X$ are permuted by a randomly generated permutation matrix $P \in \{0,1\}^{l\times l}$. Formally, the permutation prediction task is to predict the permutation matrix $P$ from a given shuffled image sequence $\tilde{X}$ such that $P^{-1} = P^T$ recovers the original ordered image sequence $X$.

We hypothesize that the learned deep models that are trained to solve this task are able to capture high-level semantic concepts, structure, and shared patterns in visual data. The ability to learn these concepts is important to perform well on the permutation prediction task, as well as to solve many other computer vision problems. Therefore, we posit that the features learned by our models are transferable to other related computer vision tasks as well.

\subsection{Learning}
Let us define a training set $\mathcal{D} = \lbrace \left(X, P\right)\mid X \in \mathcal{S}^c \text{ and } \forall P \in \mathcal{P}^l \rbrace$ composed by tuples of ordered image sequences $X$ and permutation matrices $P$. Here, $\mathcal{S}^c$ represents a dataset of ordered image sequences, orderings implied by a predetermined criterion $c$. Each $X\in\mathcal{S}^c$ is composed of $X = \left< I_1, I_2, \ldots, I_l \right>$, an ordered sequence of images $I_i$. The notation $\mathcal{P}^l$ represents the set of all $l\times l$ permutation matrices. Accordingly, the training set $\mathcal{D}$ is thus composed of all shufflings of each $X$ by all $P$. Note that given an ordered $X$, such a dataset can be generated on-the-fly by randomly permuting the order, and the size of such permuted sets scales factorially on the sequence length $l$, providing a huge amount of data with low processing and storage cost to train high capacity models. 

A permutation matrix is a binary square matrix that has exactly a single unit value in every row and column, and zeros everywhere else. Thus, these matrices form discrete points in the Euclidean space. While, permutation matrices are central to our formulations, directly working with them for deriving gradient-based optimization solvers is difficult as such solvers often start with an initial point and iteratively refines it using small (stochastic updates along gradient directions) towards an optimum. In this respect, working directly with discrete permutation matrices is not feasible. Thus, in this paper, we propose to approximate inference over permutation matrices to inference over their nearest convex surrogate, the doubly-stochastic matrices.

A double stochastic matrix (DSM) is a square matrix of non-negative real numbers with the property that every row and every column sums to one. According to the Birkhoff-von Neumann theorem~\citep{Birkhoff:1946,Von:1953}, the Birkhoff polytope $\mathcal{B}^l$ (which is the set of $l \times l$ doubly-stochastic matrices), forms a convex hull for the set of $l \times l$ permutation matrices. Consequently, it is natural to think of DSMs as relaxations of permutation matrices.

Following these ideas, we propose to learn a parametrized function $f_\theta: \mathcal{S}^c \rightarrow \mathcal{B}^l$ that maps a fixed length image sequence (of length $l$) denoted by $\tilde{X}$ to an $l \times l$ doubly stochastic matrix $Q$. In the ideal case, the matrix $Q$ should be equal to $P$. Then, our permutation learning problem can be described as, 
\begin{equation}
\begin{aligned}
& \underset{\theta}{\text{minimize}}&  & \sum_{\left(X, P\right) \in \mathcal{D}} \Delta\left(P, f_\theta(\tilde{X})\right) + R\left(\theta\right)
\end{aligned}
\label{eq:perm_learn}
\end{equation}
where $\tilde{X}$ is the image sequence $X$ permuted by the permutation matrix $P$, $\Delta(\cdot,\cdot)$ is a loss function, $\theta$ captures the parameters of the permutation learning function, and $R(\theta)$ regularizers these parameters to avoid overfitting.

\subsection{Model}
\begin{figure*}[ht]
	\begin{center}		
		\includegraphics[width=\textwidth]{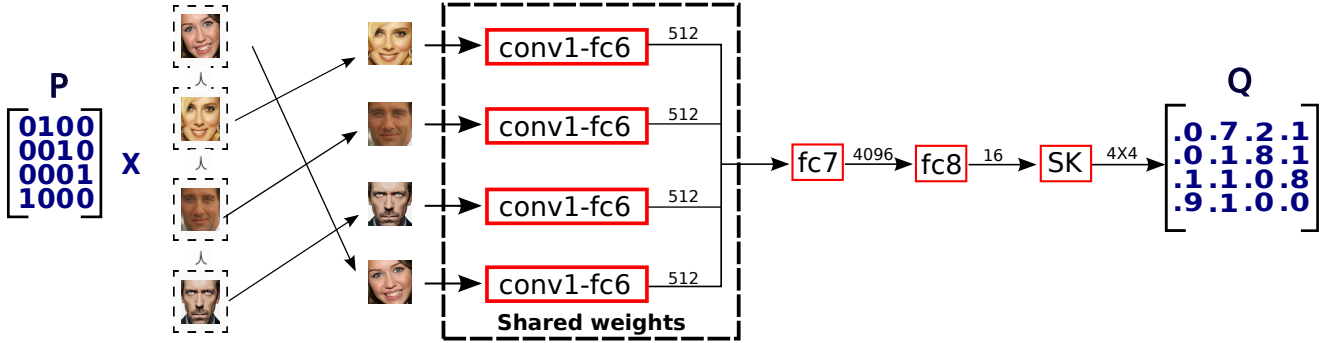}			
	\end{center}
	\caption{DeepPermNet Architecture. It receives a permuted sequence of images as input. Each image in the sequence goes trough a different branch that follows the AlexNet \protect\citep{Krizhevsky:NIPS12} architecture from \textit{conv1} up to \textit{fc6}. Then, the outputs of \textit{fc6} are concatenated and passed as input to \textit{fc7}. Finally, the model predictions are obtained by applying the Sinkhorn Layer on the outputs of \textit{fc8} layer.}
	\label{fig:cnn_model}
\end{figure*}

We wish to learn the image representation that captures the structure behind our sequences and also solves the permutation problem jointly. Then, the parametrized function $f_\theta(\cdot)$ should learn intermediate feature representations which encode semantic concepts about the input data. We propose to implement the function $f_\theta(\cdot)$ as a convolutional neural network (CNN) which is able to exploit large datasets and learn valuable low-level, mid-level, and high-level features, that can be used as intermediate representations, while jointly learning the required mapping.

More specifically, we use a Siamese type of convolutional neural network in which each branch receives an image from a permuted sequence $\tilde{X}$ (see \figref{fig:cnn_model}). Each branch up to the first fully connected layer \textit{fc6} uses the AlexNet architecture \citep{Krizhevsky:NIPS12}. The outputs of \textit{fc6} layers are concatenated and given as input to \textit{fc7}. All layers up to \textit{fc6} share the same set of weights. We name our proposed model as \textbf{DeepPermNet}.

Note that, if we ignore the structure of permutation matrices, this problem can be na\"{i}vely cast as an $l^2$ multi-label classification problem by optimizing the combination of sigmoid outputs and cross-entropy loss.  However, incorporating this inherent structure can avoid the optimizer from searching over impossible solutions, thereby leading to faster convergence and better solutions.
Thus, in the sequel, we explore schemes that uses the geometry of permutation matrices (using doubly-stochastic approximations). To the best of our knowledge, currently, there is no standard CNN layer that is able to explore such structure.

\subsubsection{Sinkhorn Normalization}
A principled and efficient way to enforce a CNN to generate DSMs as outputs is to make use of the \textit{Sinkhorn normalization} algorithm \cite{sinkhorn1964,Sinkhorn:1967}. As alluded to earlier, a DSM is a square matrix with rows and columns summing to one. Sinkhorn~\citep{sinkhorn1964,Sinkhorn:1967} showed that any non-negative square matrix (with full support~\cite{Knight:2008}) can be converted to a DSM by alternating between rescaling its rows and columns to one. Recently, \citet{Adams:2011} examine the use of DSMs as differentiable relaxations of permutation matrices in gradient based optimization problems. Here, we propose a CNN layer that performs such a normalization. Consider a matrix $Q \in \mathbb{R}_+^{l \times l}$, which can be converted to a doubly stochastic matrix by repeatedly performing row and column normalizations. Define row $R\left(\cdot\right)$ and column $C\left(\cdot\right)$ normalizations as follows,
\begin{equation}
\begin{aligned}
& R_{i, j}\left(Q\right) = \frac{Q_{i,j}}{\sum_{k=1}^{l} Q_{i,k}}; &  & C_{i, j}\left(Q\right) = \frac{Q_{i,j}}{\sum_{k=1}^{l} Q_{k,j}}\\
\end{aligned}
\end{equation}
Then, the Sinkhorn normalization for the $n$-th iteration can be defined recursively as:
\begin{align}
S^n(Q) &= \begin{cases}
       Q, & \text{if $n = 0$}\\
       C\left(R\left(S^{n-1}\left(Q\right)\right)\right), & \text{otherwise.}
\end{cases}
\end{align}

The Sinkhorn normalization function $S^n\func$ is differentiable and we can compute its gradient with respect the inputs efficiently by unrolling the normalization operations and propagating the gradient through the sequence of row and columns normalizations. For instance, the partial derivatives of the row normalizations can be defined as,
\begin{align}
\frac{\partial \Delta}{\partial Q_{p,q}} =  \sum_{j=1}^{l} \frac{\partial \Delta}{\partial R_{p,j}} \Bigg[ \frac{\ind{j=q}}{\sum_{k=1}^{l} Q_{p,k}} - \frac{Q_{p,j}}{\left(\sum_{k=1}^{l} Q_{p,k}\right)^2} \Bigg]
\end{align}
where $Q$ and $R$ are the inputs and outputs of the row normalization function and $\ind{\cdot}$ is the indicator function, evaluating to one, if its argument is true, and zero otherwise. The derivative of the column normalization can be obtained by transposing indexes appropriately. In practice, before applying the Sinkhorn normalization, we add a small value $(\approx 10^{-3})$ to each entry of $Q$ as a regularization term to avoid numerical instability.

\subsection{Inference}
As alluded to above, our main goal is to recover the original image sequence from a permuted sequence. Thus, our inference consists of approximating the closest permutation matrix $\hat{P}$ from the predicted doubly stochastic matrix $Q$. This problem can be described as,
\begin{equation}
\begin{aligned}
\hat{P} & \in \underset{\hat{P}}{\text{argmin}}&  & \norm{\hat{P} - Q}_F\\
& \text{subject to} & & \hat{P}\cdot \ones = \ones \\
& & & \ones^T \cdot \hat{P} = \ones \\
& & & \hat{P} \in \lbrace 0,1 \rbrace^{l \times l}
\end{aligned}
\end{equation}
where $\hat{P}$ is our approximated permutation matrix from $Q$. 

This optimization problem is an instance of a mixed-boolean program and can be efficiently solved by branch-and-bound methods available in public solvers \citep{cvxpy}. After obtaining $\hat{P}$ we recover the original sequence $X$ as,
\begin{align}
X = \hat{P}^T \tilde{X}. 
\end{align}

\subsection{Implementation Details}
For training, we use stochastic gradient descent with mini-batches of size 32, images of $256 \times 256$ pixels and different sequence lengths. During preprocessing, we subtract the mean and randomly crop each image to size $227 \times 227$. We initialize our network from \textit{conv1} to \textit{fc6} layers using an AlexNet model pre-trained on ILSVRC 2012 \cite{Krizhevsky:NIPS12} dataset for the task of image classification, while other layers are randomly initialized from a Gaussian distribution. Then, we set the learning rate to $10^{-5}$ and fine-tune our model for permutation prediction during 25k iterations using the multi-class cross entropy loss.

\section{Experiments}
We now describe how to extend our model to tackle different computer vision problems and measure our model performance on well established benchmarks. First, in Section~\ref{sec:perm_pred_res}, we analyze how effectively our proposed model solves the permutation prediction problem under different settings. Second, in Section~\ref{sec:relatt_res}, we evaluate our model on the relative attributes task. Finally, in Section~\ref{sec:pret_res}, we evaluate our method for self-supervised representation learning. 

\subsection{Permutation Prediction \label{sec:perm_pred_res}}
\begin{figure*}[ht]
	\begin{center}		
		\includegraphics[width=0.24\textwidth]{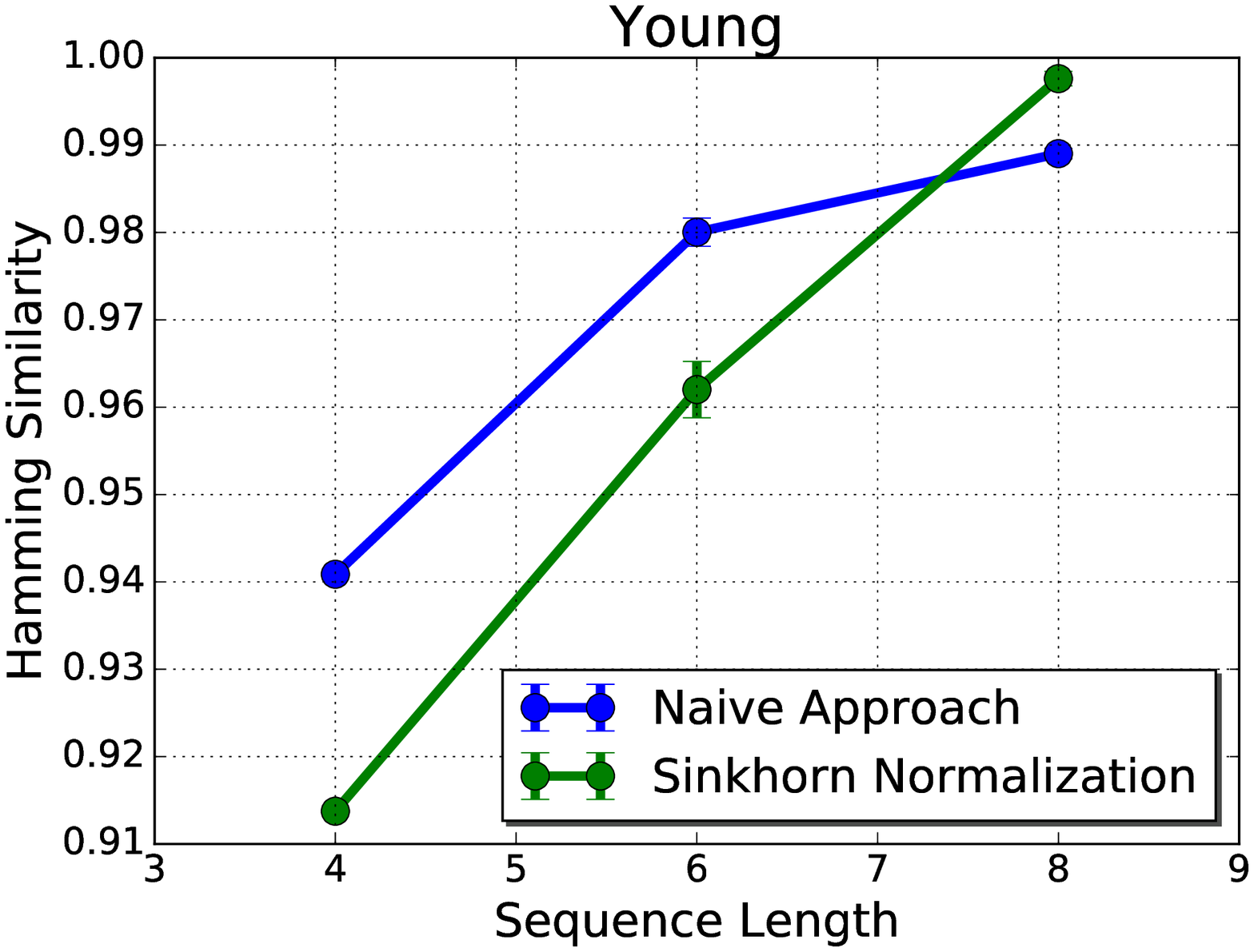}		
		\includegraphics[width=0.24\textwidth]{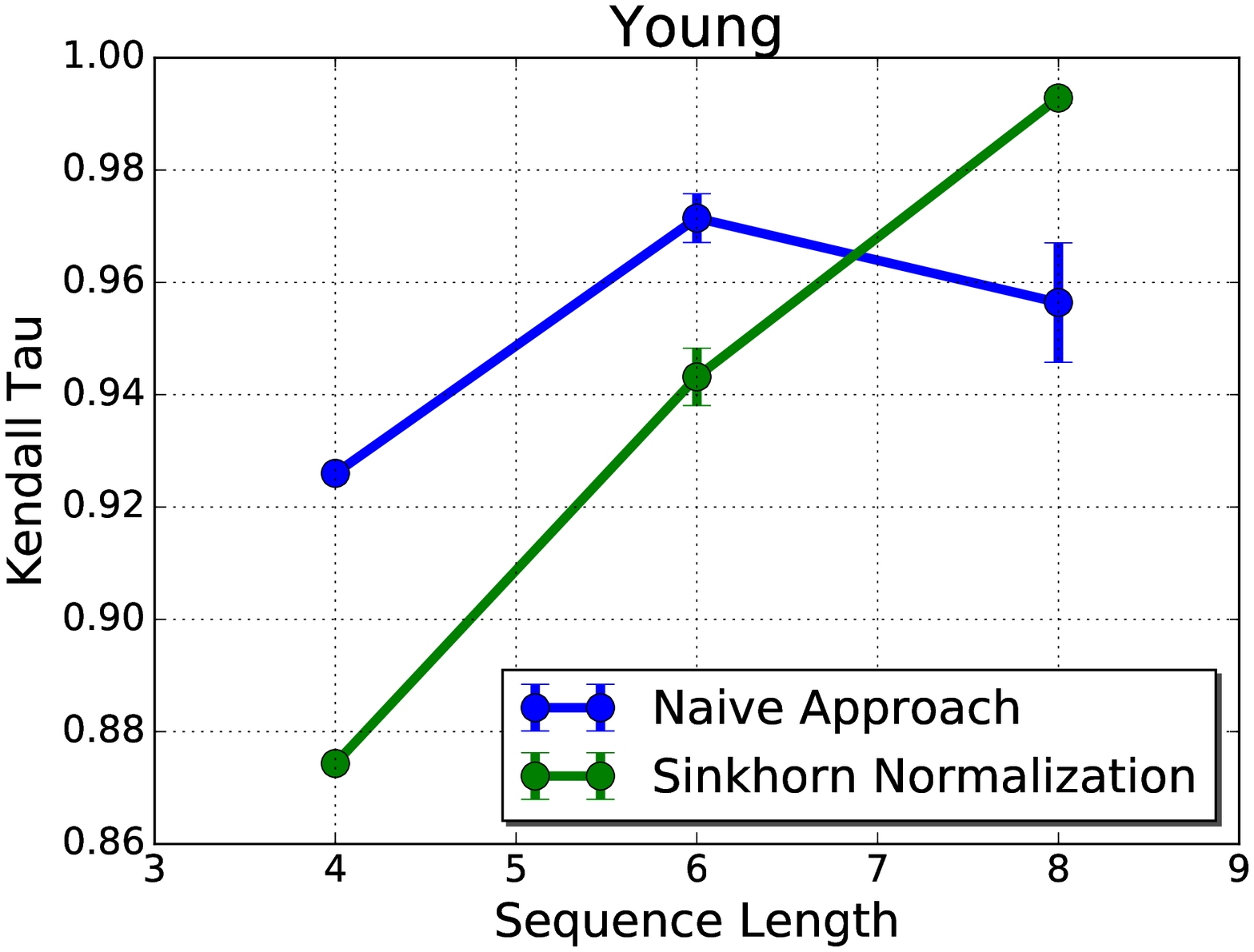}
		\includegraphics[width=0.24\textwidth]{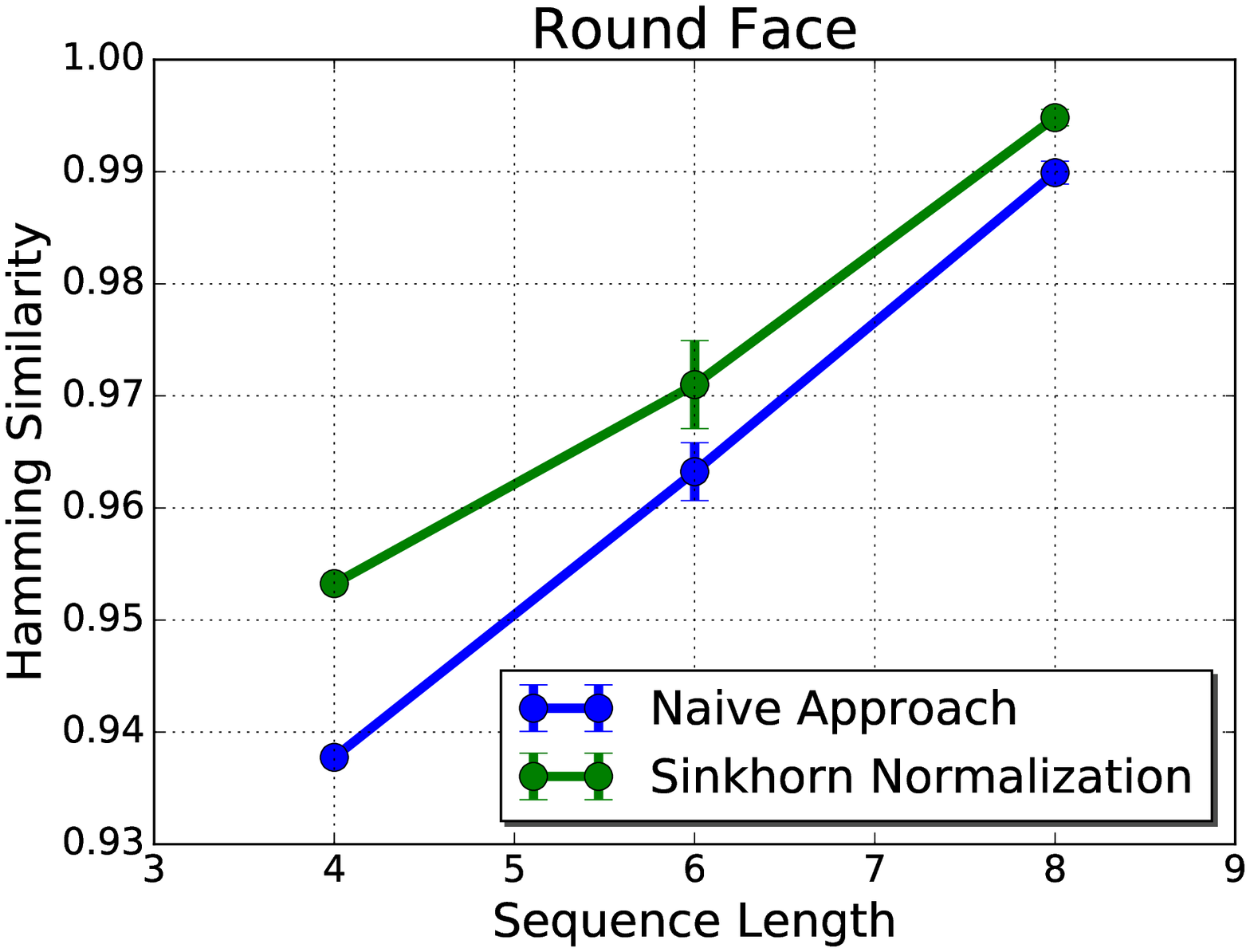}
		\includegraphics[width=0.24\textwidth]{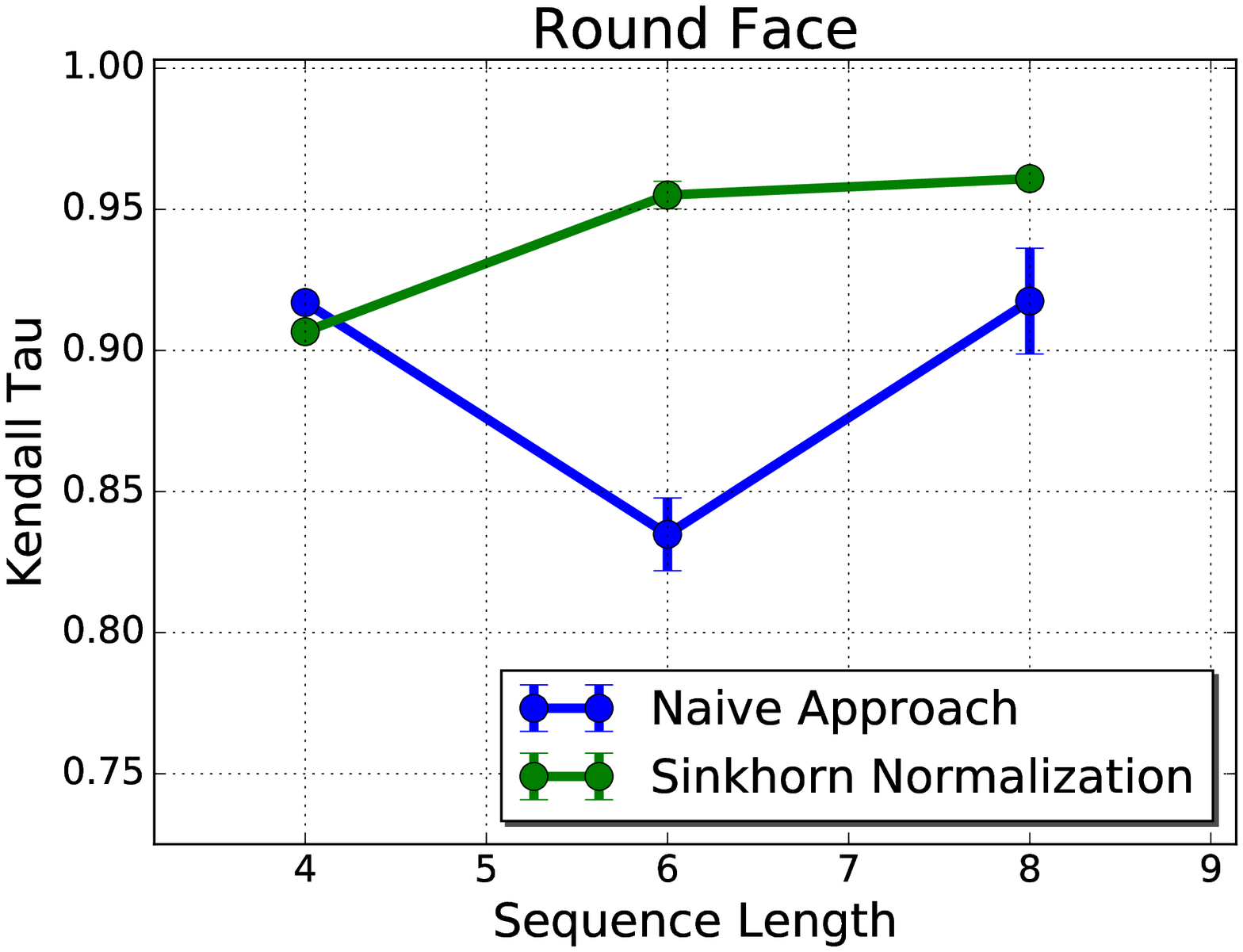}				
	\end{center}
    \vspace{-5pt}
	\caption{Evaluating the performance and stability of Sinkhorn normalization and the na\"{i}ve approach on the permutation prediction task for two attributes of public figures dataset: young and round face. Performance is reported in terms of \textit{Kendall Tau} and hamming similarity. Best viewed in color. }
	\label{fig:predperm_stab}
\end{figure*}

In this experiment, we evaluate our proposed method on the permutation prediction task and compare with a na\"{i}ve approach which combines sigmoid outputs and cross-entropy loss by casting the permutation prediction problem as a multi-label classification problem. 

In this experiment, we use the Public Figures dataset \cite{Parikh:CVPR11} which consists of 800 facial images of eight public figures and eleven physical attributes, such as big lips, white, and young. This dataset is annotated in category level, i.e., all images in a specific category may be ranked higher, equal, or lower than all images in another category, with respect to an attribute. That is, images are partially ordered.

As performance metrics for the permutation prediction task, we use Kendall-Tau and Hamming Similarity. Kendall Tau is defined as $KT=\frac{c^+-c^-}{0.5l(l-1)}$, where $c^+$ and $c^-$ denote the number of all pairs in the sequence that are correctly and incorrectly ordered, respectively. It captures how close we are to the perfect rank. The Hamming similarity measures the number of equals entries in two vectors or matrices normalized by the total number of elements. It indicates how similar our prediction is to the ground truth permutation matrix. In addition, we measure the averaged $\ell_1$ normalization error of rows and columns of our predicted permutation matrices.

We train a CNN model for each attribute in the Public Figures dataset by sampling 30K ordered image sequences from as training images. Then, we evaluate the trained models on 20K image sequences generated from the test set by sampling correctly ordered sequences and randomly permuting them. We averaged the results over the 11 attributes and repeat the experiment for image sequences of length 4 and 8. \tablename{~\ref{table:perm_pred}} presents the results for Sinkhorn Normalization and the na\"{i}ve approach. 
\begin{table}[t]
\centering
\begin{tabular}{l|c|c|c|c}
\textbf{Method} & \textbf{Length} & KT & HS & NE\\
\hline
Naive App. & 4 & 0.859 & 0.893  & 0.062  \\
Sinkhorn Norm. & 4 & \textbf{0.884} & \textbf{0.906}  & \textbf{0.019} \\ \hline
Naive App. & 8 & 0.774  & 0.832 & 0.1  \\ 
 Sinkhorn Norm.& 8 & \textbf{0.963} & \textbf{0.973}  & \textbf{0.022} \\
\hline
\end{tabular}
\vspace{5pt}
\caption{Evaluating and comparing Sinkhorn Normalization and a combination of sigmoid and cross entropy loss, named naive approach, on the permutation prediction task using the Public Figures Dataset \protect\cite{Parikh:CVPR11}. KT, HS and NE stands for \textit{Kendall Tau}, hamming similarity and $l_1$ normalization error respectively.}
\label{table:perm_pred}
\end{table}

We observe the na\"{i}ve approach works well for small sequences and is able to learn the normalization by itself. As the sequence length increases, however, the performance of the na\"{i}ve approach degenerates and the $\ell_1$ normalization error increases. On the other hand, the Sinkhorn Normalization approach reaches better results in \textit{Kendall-Tau} and hamming similarity while keep the normalization error almost unchangeable even for longer sequences. This fact suggests that exploring the geometrical structure of the space of doubly-stochastic matrices (and thereby the permutation matrices) is useful.

The number of possible permutations for a given image sequence increases factorially with the sequence length. Thus, we evaluate how stable are the predictions of our proposed model in relation to the variation in permutations. For a given attribute, we create a test dataset composed of 1K random image sequences of length four, six, and eight. Then, we augment this dataset with as many permutations as possible for each sequence. More specifically, we generated all permutations of length four for each test sample. Due to computational complexity, for length six and eight we randomly sampled fifty different permutations for each test sample. \figurename{~\ref{fig:predperm_stab}} shows the results for the attributes Young and Round Face. 

Again, we obtain better results with the Sinkhorn Normalization layer which shows a trend of increasing performance as the length of the image sequence increases. We also note that the variations in the prediction for the Sinkhorn Normalization layer is negligible while the na\"{i}ve approach becomes unstable for longer sequences. Therefore, we adopt the sinkhorn normalization layer in our proposed model DeepPermNet. In the next sections, we focus on how the DeepPermNet can be applied for ordering images based on relative attributes and self-supervised representation learning.

\subsection{Relative Attributes \label{sec:relatt_res}}
In this experiment, we use DeepPermNet to compare images in a given  sequence according to a certain attribute by predicting permutations and applying its inverse. This procedure can be used to solve the relative attributes task, the goal of which is to compare pairs or sets of images according to the ``strength'' of a given attribute.

For this application, we use the OSR scene dataset \citep{Parikh:CVPR11}, in addition to the Public Figures Dataset\cite{Parikh:CVPR11}. We train our model for each attribute with 30K ordered image sequences of length 8 generated from the training set. Then, we report our model performance in terms of pairwise accuracy measured on the predicted ordering for 20K image sequences of length 8 generated from the test set using stratified sampling.
\begin{table*}[t]
\centering
\caption{Evaluating relative attributes prediction model on Public Figures Dataset.}
 \resizebox{\textwidth}{!}{\begin{tabular}{l|c|c|c|c|c|c|c|c|c|c|c|c}
\textbf{Method} &  \textbf{Lips} & \textbf{Eyebrows} & \textbf{Chubby} & \textbf{Male} & \textbf{Eyes} & \textbf{Nose} & \textbf{Face} & \textbf{Smiling} & \textbf{Forehead} 	& \textbf{White} & \textbf{Young} & \textbf{Mean} \\
\hline
\citet{Parikh:CVPR11} & 79.17 & 79.87 & 76.27 & 81.80 & 81.67 & 77.40 & 82.33 & 79.90 & 87.60  & 76.97 & 83.20 & 80.56\\
\citet{LI:ACCV12} &81.87 & 81.84 & 79.97 & 85.33 & 83.15 & 80.43 & 86.31 & 83.36 & 88.83  & 82.59 & 84.41 & 83.37\\
\citet{Yu:CVPR14} & 90.43 & 89.83 & 87.37 & 91.77 & 91.40 & 89.07 & 86.70 & 87.00 & 94.00  & 87.43 & 91.87 & 89.72\\
\citet{Souri:ACCV16} & 93.62 & 94.53 & 92.32 & 95.50 & 93.19 & 94.24 & 94.76 & 95.36 & 97.28  & 94.60 & 94.33 & 94.52\\
\hline
DeepPermNet & \textbf{99.55} &\textbf{ 97.21} & \textbf{97.66} & \textbf{99.44} & \textbf{96.54} & \textbf{96.21} & \textbf{99.11} & \textbf{97.88} & \textbf{99.00}  & \textbf{97.99} & \textbf{99.00} & \textbf{98.14}\\
\hline
\end{tabular}}
\label{table:rel_att_pubfig}
\end{table*}
\begin{table*}[t]
\centering
\caption{Evaluating relative attributes prediction methods on OSR dataset.}
\begin{tabular}{l|c|c|c|c|c|c|c}
\textbf{Method} & \textbf{Depth-Close} & \textbf{Diagonal-Plane} & \textbf{Natural} & \textbf{Open} & \textbf{Perspective} & \textbf{Size-Large} & \textbf{Mean}  \\
\hline
\citet{Parikh:CVPR11} & 87.53 & 86.5  & 95.03 & 90.77 & 86.73 & 86.23 & 88.80 \\
\citet{LI:ACCV12} & 89.54 & 89.34 & 95.24 & 92.39 & 87.58 & 88.34 & 90.41 \\
\citet{Yu:CVPR14} & 90.47 & 92.43 & 95.7  & 94.1  & 90.43 & 91.1 & 92.37 \\
\citet{Singh:ECCV2016} & 96.1  & 97.64 & 98.89 & 97.2  & 96.31 & 95.98 & 97.02\\
\citet{Souri:ACCV16} & 97.65 & \textbf{98.43} & \textbf{99.4}  & 97.44 & 96.88 & 96.79 & 97.77 \\
\hline
DeepPermNet (AlexNet) & 96.09 & 94.53 & 97.21 & 96.65 & 96.46 & 98.77 & 96.62 \\
DeepPermNet (VGG16) & \textbf{96.87} & 97.99 & 96.87 & \textbf{99.79} & \textbf{99.82} & \textbf{99.55} & \textbf{98.48} \\
\hline
\end{tabular}
\label{table:rel_att_osr}
\end{table*}

Different from the existing methods \citep{Souri:ACCV16,Singh:ECCV2016}, we directly predict the order for sequences of images instead of pairs. Our scheme allows us to make use of the structure in the sequences as a whole, which is more informative than pairs. For a fair comparison to prior methods, we measure our performance by computing the pairwise accuracy for all pairs in each sequence. Tables \ref{table:rel_att_pubfig} and \ref{table:rel_att_osr} present our results.

On Public Figures dataset, DeepPermNet outperforms the state-of-the-art models by a margin of 3\% in pairwise accuracy. It is a substantial margin, consistently observed across all attributes. Note that, we outperform the recent method in~\citep{Souri:ACCV16}, which is a VGG CNN model that has significantly more modeling capacity than the AlexNet \citep{Krizhevsky:NIPS12} architecture we use. On the other hand, our method works slightly worse than~\citep{Souri:ACCV16} on OSR dataset. We also provide results by building our scheme on a VGG CNN model. As is clear, using this variant, we demonstrate even better results outperforming the state-of-the-art methods.

It is worth noting that DeepPermNet works better when we use longer sequences for training, because they provide rich information, which can be directly used in our method. For instance, the performance of DeepPermNet drops 7\% in terms of average pairwise accuracy on Public Figures dataset when we train our model using just pairs. In addition, the proposed model is not able to explicitly handle equality cases, since the permutation learning formulation assumes each permutation is unique, which is not true in the relative attributes task. Perhaps, this is the reason for the difference in performance between Public Figures and OSR datasets. Nonetheless, DeepPermNet is able to learn very good attribute rankers from data as shown in our experiments.

\begin{figure}[b]
	\begin{center}		
		\includegraphics[width=0.5\textwidth]{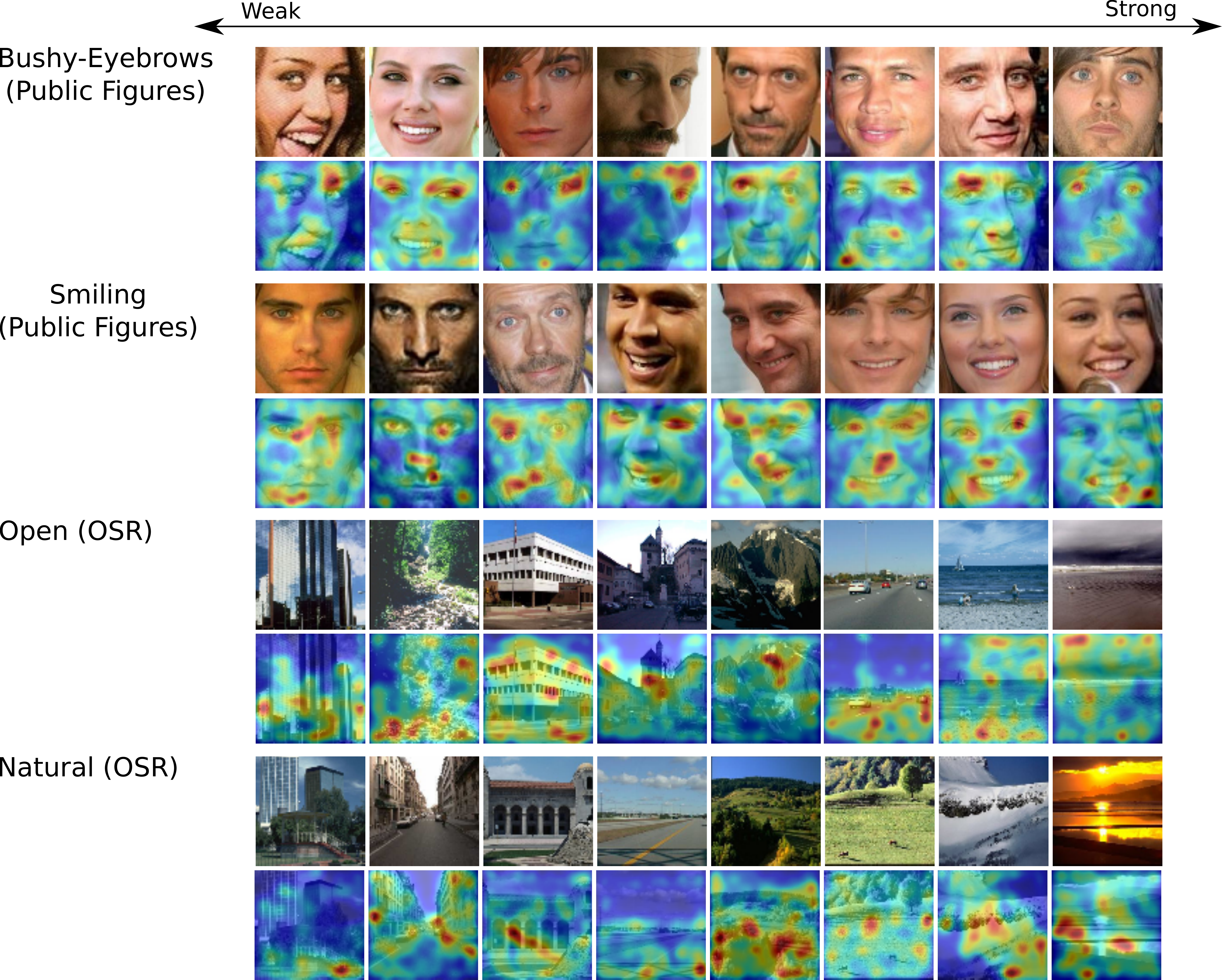}		
	\end{center}    
	\caption{Qualitative results on Public Figures and OSR test images. Better viewed in color.}
	\label{fig:relatt_demo}
\end{figure}

We also compute the saliency maps of different attributes using the method proposed by \citet{Simonyan:2013}. More specifically, we take the derivative of the estimated permutation matrix with respect to the input, given a set of images. We perform max pooling across channels to generate the saliency maps. \figref{fig:relatt_demo} presents qualitative results and saliency maps generated by DeepPermNet for different attributes.

These maps are a simplified way to visualize which pixels, regions, and features of a given image are more relevant to the respective permutations predicted by our method. For instance, the attribute ``bushy eyebrows" are sensitive to the region of eyes, while the attribute ``smiling'' is more sensitive to the mouth region. An interesting observation is the possibility of localizing such features without any explicit supervision (e.g., bounding boxes or segmentation masks), which could be used for unsupervised attribute localization.

\subsection{Self-Supervised Representation Learning \label{sec:pret_res}}
\citet{Yosinski:2014} observed that pre-training a network helps to regularize the model reaching better performance on the test set. Motivated by this observation, we propose to use the spatial structure existent in images as self-supervisory signal to generate ordered sequences of patches to train our model and transfer the learned weights for supervised target tasks such as object classification, detection, and segmentation.

More specifically, we use the train split of the ImageNet dataset \cite{Krizhevsky:NIPS12} as training set discarding its labels. For each image, we split it into a grid with $3 \times 3$ cells, extract a patch of size $64 \times 64$ pixels within each grid cell and generate a sequence where the ordering is established by the spatial position of each patch in the grid (see Figure~\ref{fig:perm_intro} right). Then, we train our models to predict random permutations of these generated patch sequences as before. 

Unlike our previous experiments, we train our CNN models from scratch starting using random initialization. The model is trained for 400k iterations using a initial learning rate of $0.001$, which is dropped to one-tenth every 100k iterations. we use batches of 256 sequences each of $64 \times 64$  image patches. Note that as this is self-supervised training, we do not use any pre-trained models or human labels.

Using a CNN to recover an image from its parts is a challenging task because it requires the network to learn semantic concepts, contextual information, and objects-parts relationships, in order to predict the right permutation. In order to evaluate how well the proposed models can solve such a task, we use 50k images on the ImageNet validation set and apply random permutations using the $3 \times 3$ grid layout. In this self-supervised setting, on the Kendall Tau metric, DeepPermNet reaches a score of $0.72$ and the na\"{i}ve approach reaches $0.64$ on the permutation prediction task.

\begin{table}[t]
\centering
 \resizebox{0.48\textwidth}{!}{\begin{tabular}{l|c|c|c}
\textbf{Pre-training Method} & \thead{Classification\\(mAP\%)} & \thead{FRCN Detection \citep{Girshick:ICCV15}\\(mAP\%)} & \thead{FCN Segmentation \citep{Long:CVPR15}\\(\%mIU)}\\
\hline
ImageNet & 78.2 & 56.8 & 48.0 \\
Random Gaussian & 53.3 & 43.4 & 19.8\\
\hline
\citet{Agarwala:2004} & 52.9 & 41.8 & - \\
\citet{Doersch:ICCV15} & 55.3 & 46.6 & - \\
\citet{Wang:ICCV15} & 58.4 & 44.0 & -\\
\citet{Pathak:2016} & 56.5 & 44.5 & 29.7 \\
\citet{Donahue:2016} & 58.9 & 45.7 & 34.9 \\
\citet{Zhang:ECCV2016} & 65.6 & 47.9 & 35.6 \\
\citet{Noroozi:ECCV16} & 68.6 & \textbf{51.8} & 36.1 \\
\hline
Naive Approach & 65.9 & 48.2 & 36.8\\
DeepPermNet & \textbf{69.4} & 49.5 & \textbf{37.9}\\
\hline
\end{tabular}}
\vspace{2pt}
\caption{Classification and detection results on PASCAL VOC 2007 test set under the standard mean average precision (mAP), and segmentation results on the PASCAL VOC 2012 validation set under mean intersection over union (mIU) metric.}
\label{table:pretraining}
\end{table}
Following the literature on self-supervised pre-training \citep{Doersch:ICCV15,Donahue:2016,Pathak:2016,Noroozi:ECCV16}, we test our models on the commonly used self-supervised benchmarks on the PASCAL Visual Object Challenge and compare against supervised and self-supervised procedures for pre-training. We transfer our learned weights to initialize from \textit{Conv1} to \textit{Conv5} layers of AlexNet \citep{Krizhevsky:NIPS12}, Fast-RCNN \citep{Girshick:ICCV15} and Fully Convolutional Network \citep{Long:CVPR15} models and fine-tune them for object classification, object detection, and object segmentation tasks respectively, using their default training parameters. For object classification and detection, we report the mean average precision (mAP) on PASCAL VOC 2007 \cite{PASCAL:VOC2007}, while for object segmentation, we report mean average intersection over union (mIU) on PASCAL VOC 2012 \citep{PASCAL:VOC2012}. \tabref{table:pretraining} presents our results.

\begin{figure}[t]
	\begin{center}		
        \includegraphics[width=0.23\textwidth]{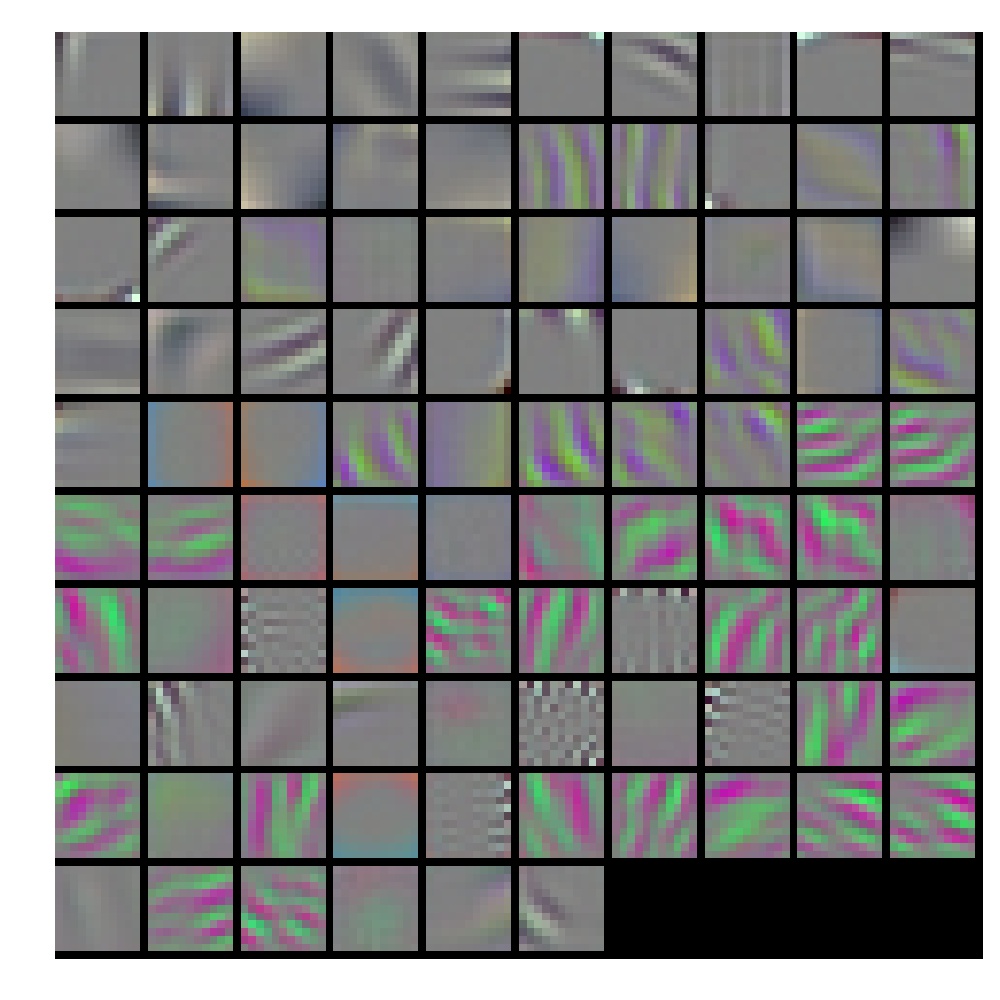}			
		\includegraphics[width=0.23\textwidth]{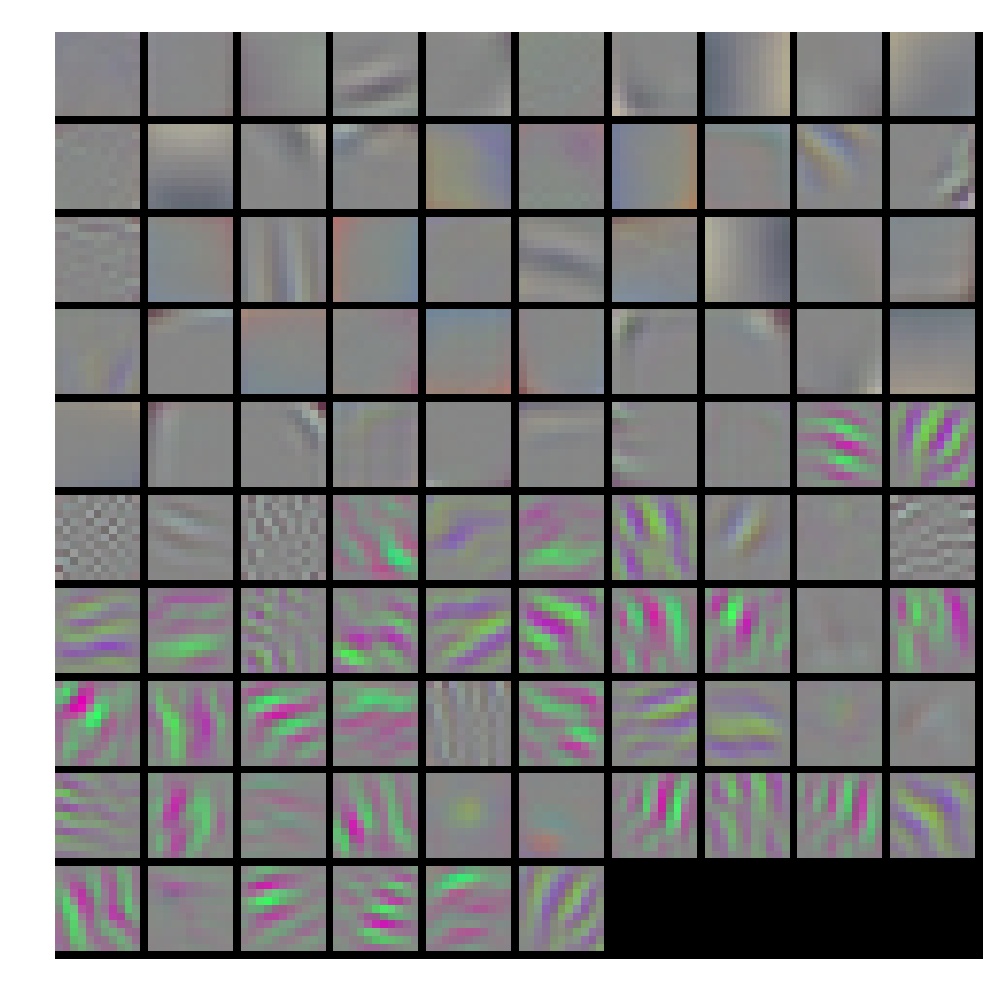}
	\end{center}	
	\caption{Comparison of convolution layer one filters for left \protect\citet{Noroozi:ECCV16} and right our method.}
	\label{fig:unsup_filters}
\end{figure}

We observe that the self-supervised methods are still behind the supervised approach, but this gap reduces gradually. Our DeepPermNet outperforms the self-supervised competitors in object classification and segmentation, while it produces the second best performance for the detection task. Interestingly, when finer grid cells are used (e.g., $4\times4$), we do not observe any improvement in recognition performance. Moreover, the first layer filters learned by our method and the ones learned by \citet{Noroozi:ECCV16} seems somewhat similar (see \figref{fig:unsup_filters}) despite the different number of permutations learned. This fact suggests that $3\times3$ grid partition and a well chosen subset of permutations are enough to learn filters which produce state-of-the-art results for self-supervised representation learning. However, DeepPermNet is a generic method than the method proposed by \citet{Noroozi:ECCV16}, and our method can be used to solve many different computer vision tasks as shown in our experiments.

\section{Conclusion}
In this paper, we tackled the problem of learning the structure of visual data by introducing the task of visual permutation learning.
We formulated an optimization problem for this task with the goal of recovering the permutation matrix responsible for generating a given randomly shuffled image sequence based on a pre-defined visual criteria. We proposed a novel CNN layer that can convert standard CNN predictions to doubly-stochastic approximations of permutation matrices using Sinkhorn normalizations; this CNN can be trained in an end-to-end manner. Through a variety of experiments, we showed that our framework is generic and could be used not only for recovering the order, but also to generate good initializations for training standard CNN models. 

Going forward, one compelling direction of investigation is to replace our unrolling of Sinkhorn iterations for gradient propagation by more standard and exact optimizers. Another important direction is to investigate the use of this scheme for other modalities of data such as video sequences and 3D data.

\small{
\smallskip \noindent \textbf{Acknowledgements:} This research was supported by the Australian Research Council (ARC) through the Centre of
Excellence for Robotic Vision (CE140100016) and was undertaken with the resources from the National Computational Infrastructure (NCI), at the Australian National University (ANU).}

{\bibliography{short,unsuprep_bib}}

\end{document}